\documentclass[sigconf]{acmart}
\AtBeginDocument{%
  }

\copyrightyear{2026}
\acmYear{2026}
\setcopyright{cc}
\setcctype{by}
\acmConference[MM '26]{Proceedings of the 35th ACM International Conference on Multimedia}{November 10--14, 2026}{Rio de Janeiro, Brazil}
\acmBooktitle{Proceedings of the 35th ACM International Conference on Multimedia (MM '26), November 10--14, 2026, Rio de Janeiro, Brazil}
\acmDOI{10.1145/3767308.3835011}
\acmISBN{979-8-4007-2213-4/2026/11}

\usepackage{tabularx}  
\usepackage{booktabs}  
\usepackage{balance}
\usepackage{algorithm}
\usepackage{algorithmic}
\usepackage{graphicx}
\usepackage{booktabs}
\usepackage{listings}
\usepackage{pifont}
\usepackage{multirow, makecell}

\definecolor{our_blue}{rgb}{0.85, 0.9, 0.95}
\definecolor{clip_red}{rgb}{0.961, 0.749, 0.737}  

\usepackage{array}
\usepackage{pifont}

\usepackage{hyperref}

\usepackage{bbold}
\usepackage{amsmath}  

\begin{document}

\title{DynActiveGS: Active Gaussian Splatting for Dynamic Scene Reconstruction}

%





\author{Hongbo Duan}
\affiliation{%
  \institution{Tsinghua University, Shenzhen International Graduate School, Peng Cheng Laboratory}
  \city{Shenzhen}
  \state{Guangdong}
  \country{China}
}
\email{dhb24@mails.tsinghua.edu.cn}
\orcid{0009-0002-8463-2170}

\author{Pengting Luo}
\affiliation{%
  \institution{Huawei Technologies Ltd}
  \city{Shenzhen}
  \state{Guangdong}
  \country{China}
}
\email{pengtingluo@gmail.com}

\author{Chengzhi Zhao}
\affiliation{%
  \institution{Harbin Institute of Technology, Shenzhen}
  \city{Shenzhen}
  \state{Guangdong}
  \country{China}
}
\email{2021210731@stu.hit.edu.cn}

\author{Yuanhao Chiang}
\affiliation{%
  \institution{Tsinghua University, Shenzhen International Graduate School}
  \city{Shenzhen}
  \state{Guangdong}
  \country{China}
}
\email{jiang-yh24@mails.tsinghua.edu.cn}

\author{Fangming Liu}
\affiliation{%
  \institution{Peng Cheng Laboratory}
  \city{Shenzhen}
  \state{Guangdong}
  \country{China}
}
\email{fangminghk@gmail.com}

\author{Xueqian Wang}
\authornote{Corresponding author}
\affiliation{%
  \institution{Tsinghua University, Shenzhen International Graduate School}
  \city{Shenzhen}
  \state{Guangdong}
  \country{China}
}
\email{wang.xq@sz.tsinghua.edu.cn}

%
\renewcommand{\shortauthors}{Duan et al.}

\begin{abstract}
We present DynActiveGS, a dynamic-aware active reconstruction framework based on 3D Gaussian Splatting (3DGS) for autonomous exploration in dynamic environments. The framework incrementally reconstructs a 3D Gaussian scene representation while suppressing motion-corrupted observations through online uncertainty prediction and uncertainty-weighted Gaussian optimization. A key component of DynActiveGS is the explicit decomposition of uncertainty into structural uncertainty and motion-induced uncertainty, which enables the system to distinguish under-reconstructed static regions from dynamically unreliable areas. Based on these uncertainty fields, DynActiveGS performs dynamic-aware viewpoint selection and dynamic-constrained path planning to favor informative yet stable observations during exploration. The resulting system forms a unified closed-loop pipeline for robust active reconstruction in dynamic scenes. Extensive experiments on challenging dynamic benchmarks demonstrate consistent improvements over existing active reconstruction baselines in reconstruction accuracy, completeness, rendering quality, and exploration efficiency.
\end{abstract}

%
\begin{teaserfigure}
  \includegraphics[width=\textwidth]{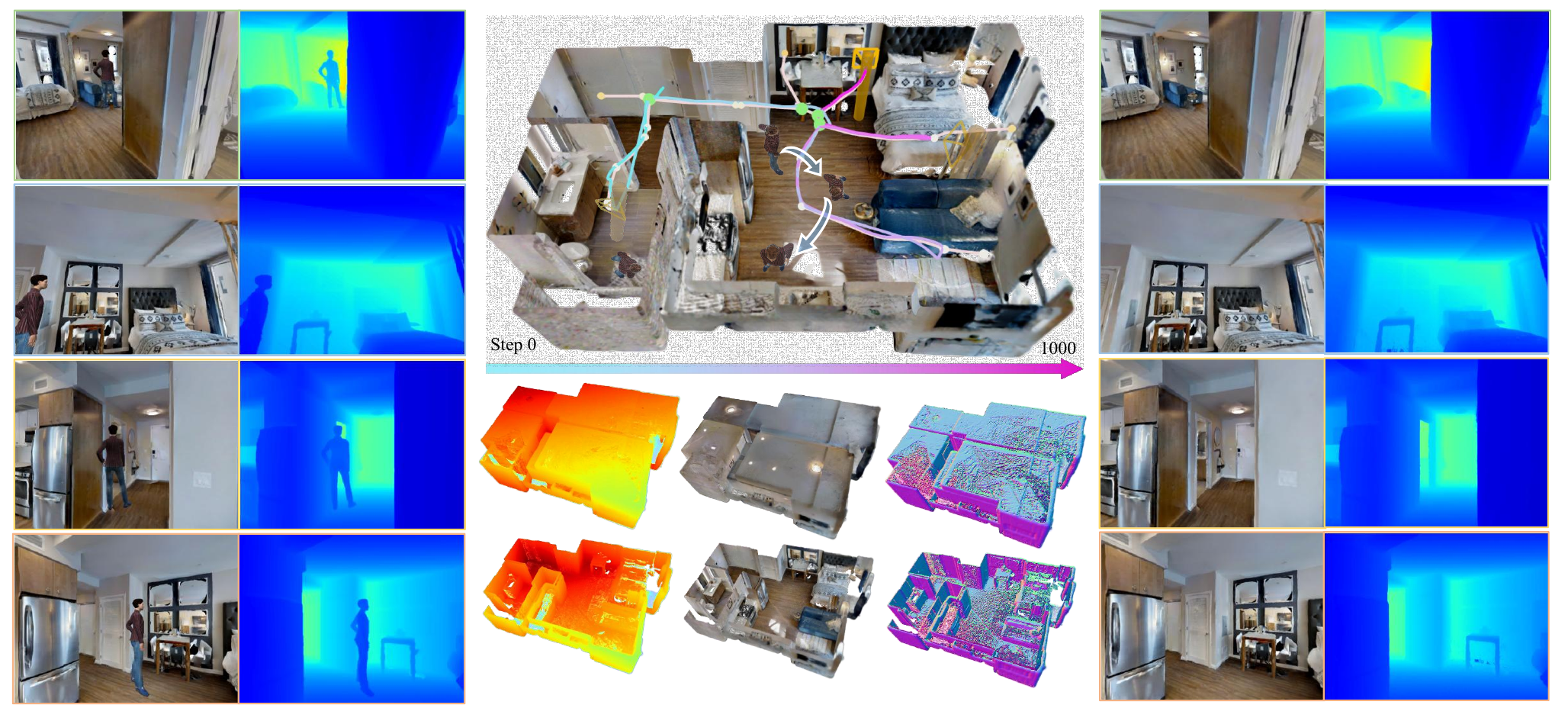}
    \caption{\textbf{DynActiveGS in action.}
    The robot actively explores a dynamic indoor scene and progressively improves reconstruction completeness.
    From left to right, we show RGB-D inputs with dynamic disturbances, intermediate reconstruction states, and final renderings after 1000 steps.
    DynActiveGS preserves stable geometry and rendering quality despite dynamic interference.}
  \label{fig:teaser}
\end{teaserfigure}


\maketitle

\section{Introduction}

The ability to reconstruct 3D scenes from visual observations is fundamental to computer vision and robotics~\cite{newcombe2011kinectfusion,whelan2015elasticfusion,dai2017bundlefusion}.
Beyond passive perception, intelligent agents should actively determine where to move and what to observe~\cite{isler2016information} to efficiently acquire informative data and build high-fidelity scene representations~\cite{kirsch2019batchbald}.
This paradigm, known as \emph{active 3D reconstruction}~\cite{aloimonos1988active}, integrates perception, uncertainty estimation, and motion planning~\cite{chen2011active}.

Recent advances in 3D Gaussian Splatting (3DGS)~\cite{kerbl20233d} have enabled explicit, differentiable, and efficient scene representations with high-quality rendering and incremental optimization.
These advantages have inspired active reconstruction methods that leverage 3DGS with next-best-view (NBV) planning~\cite{peralta2020next,pito2002solution} and uncertainty-driven exploration~\cite{lee2022uncertainty,pan2022activenerf,ran2023neurar}.
However, existing active reconstruction approaches~\cite{feng2024naruto,jin2025activegs,li_activesplat_2025,polyzos_activeinitsplat_2025,chen2025activegamer,li_active3d_2025,jin2024gs,HGS-Planner,YuhanRSS25} are primarily developed under the assumption of static environments.

Dynamic objects, including pedestrians, vehicles, and articulated agents, are common in real-world scenarios and introduce fundamental challenges to active reconstruction~\cite{huang20242d,yu2024mip}.
Motion-corrupted observations can degrade Gaussian optimization~\cite{palazzolo2019refusion,xu2024dg}, distort uncertainty estimation~\cite{ren2024nerf,kulhanek2024wildgaussians}, and mislead viewpoint planning~\cite{kuang2024active,jiang2024fisherrf}.
Meanwhile, existing dynamic 3DGS and Gaussian-based SLAM methods mainly focus on passive mapping and dynamic object filtering~\cite{xu2024dg,sandstrom2025splat,zheng2025wildgs}, leaving active perception and long-horizon exploration in dynamic environments largely unexplored.
Therefore, robust active reconstruction under dynamic scenes remains an open challenge.

In this paper, we propose \emph{DynActiveGS}, a dynamic-aware active reconstruction framework for 3D Gaussian Splatting in dynamic environments.
As shown in Fig.~\ref{fig:teaser}, DynActiveGS first performs uncertainty-aware Gaussian reconstruction by predicting pixel-wise uncertainty and suppressing motion-corrupted observations during online optimization.
Based on the reconstructed Gaussian map, we further disentangle structural uncertainty from motion-induced uncertainty to construct map-level uncertainty fields for exploration.
These uncertainty fields guide dynamic-aware viewpoint selection through local-global scoring on a Voronoi graph, while a motion-constrained planner generates efficient and stable trajectories.
Together, these components establish a unified perception--planning--reconstruction pipeline for closed-loop active reconstruction in dynamic scenes.

We evaluate DynActiveGS on diverse dynamic benchmarks.
Results demonstrate consistent improvements over existing active reconstruction baselines in reconstruction accuracy, completeness, rendering quality, and exploration efficiency.
Our work extends active 3D reconstruction beyond static environments toward robust embodied perception in dynamic real-world scenarios.

\noindent Our \textbf{contributions} are summarized as follows:
\begin{itemize}
    \item We propose DynActiveGS, a dynamic-aware active reconstruction framework for 3D Gaussian Splatting in dynamic scenes.
    
    \item We introduce an uncertainty-aware Gaussian reconstruction strategy that predicts online uncertainty, suppresses motion-corrupted observations, and disentangles structural and motion-induced uncertainty.
    
    \item We develop a closed-loop framework integrating dynamic-aware viewpoint selection and motion-constrained path planning, achieving robust active exploration in challenging dynamic reconstruction scenarios.
\end{itemize}

\section{Related Work}

\subsection{3D Scene Representation}

Scene representation plays a critical role in balancing reconstruction fidelity, efficiency, and online perception capability. Classical 3D reconstruction relies on structure-from-motion (SfM) pipelines~\cite{liu2022planemvs}, which estimate camera poses and scene geometry through feature matching~\cite{schonberger2016structure,sandstrom2025splat}, geometric verification, and bundle adjustment~\cite{chng2022gaussian}. Although effective, these methods are computationally expensive and mainly assume static environments. Neural scene representations, such as NeRFs~\cite{mildenhall2021nerf} and their variants~\cite{barron2022mip,li2022bnv,chen2022tensorf}, achieve high-quality view synthesis but remain inefficient for real-time deployment. Recently, 3D Gaussian Splatting (3DGS)~\cite{kerbl20233d} has emerged as an efficient explicit representation with real-time rendering and incremental optimization, enabling promising applications in online reconstruction and robotic perception. Recent Gaussian-based mapping and SLAM systems further demonstrate its potential, yet most remain limited to static scenes~\cite{HGS-Planner,keetha2024splatam,li2024geogaussian}.

\subsection{Active 3D Reconstruction}

Active reconstruction focuses on selecting informative viewpoints to improve scene acquisition efficiency~\cite{chen2024gennbv,li2025nextbestpath}. Early approaches exploit geometric uncertainty, occupancy reasoning, or information gain for next-best-view (NBV) planning~\cite{zhang2025peering,pan2022activenerf,jiang2024fisherrf}. Recent methods further integrate neural representations and uncertainty estimation to guide exploration~\cite{lee2022uncertainty,feng2024naruto,kuang2024active,yan_active_2023}. With the development of NeRF and 3DGS representations, several works investigate active reconstruction with neural and Gaussian models~\cite{jin2025activegs,li_activesplat_2025,jin2024gs,HGS-Planner}. However, existing frameworks are predominantly designed for static environments, where dynamic objects are ignored or treated as noise, leading to unreliable optimization and viewpoint selection.

\subsection{Dynamic Scene Reconstruction}

Dynamic scenes pose significant challenges for reliable 3D reconstruction. Traditional SLAM methods handle moving objects through motion segmentation and robust optimization~\cite{bescos2018dynaslam,jiang2024rodyn}, while recent neural and Gaussian-based approaches model or filter dynamic content for improved robustness~\cite{xu2024dg,sandstrom2025splat,zheng2025wildgs}. Although uncertainty-aware Gaussian mapping improves reconstruction under dynamic conditions, existing approaches mainly focus on passive mapping and tracking rather than active exploration~\cite{keetha2024splatam,zhu_vigs-slam_2025}. Therefore, they cannot determine how an embodied agent should actively acquire informative observations in dynamic environments. Our work bridges this gap by integrating dynamic scene modeling with active viewpoint planning in a unified Gaussian splatting framework.


\section{Method}

\begin{figure*}[t]
  \centering
  \includegraphics[width=\linewidth]{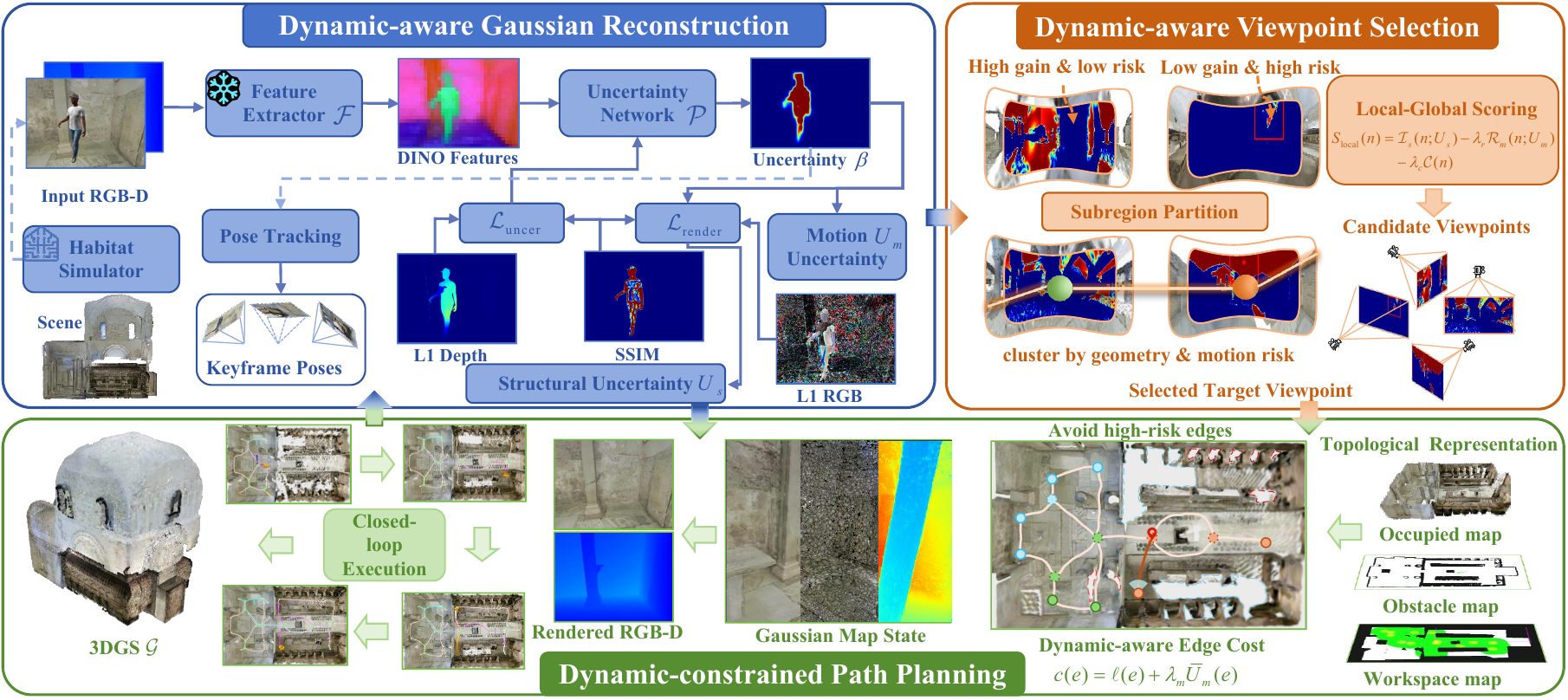}
\caption{\textbf{Overview of DynActiveGS.}
Given RGB-D observations in dynamic environments, DynActiveGS first performs dynamic-aware Gaussian reconstruction by predicting a per-pixel uncertainty map and updating the 3D Gaussian representation through uncertainty-weighted optimization. It then constructs structural uncertainty $U_s$ and motion uncertainty $U_m$ to guide dynamic-aware viewpoint selection via local-global scoring on a Voronoi graph. Finally, a dynamic-constrained path planner computes stable trajectories using motion-aware edge costs, enabling closed-loop active reconstruction under dynamic disturbances.}
  \label{fig:DynActiveGS_overview}
\end{figure*}

\label{sec:method}

As shown in Fig.~\ref{fig:DynActiveGS_overview}, DynActiveGS operates as a closed-loop active reconstruction system in dynamic environments. At each timestep $t$, the embodied agent acquires an RGB-D observation $(I_t, D_t)$ and an estimated camera pose $\mathbf{T}_t \in SE(3)$. The system updates the 3D Gaussian scene representation through uncertainty-aware reconstruction, constructs structural uncertainty $U_s$ and motion uncertainty $U_m$, selects the next informative viewpoint, and executes a dynamically stable path toward it. The loop continues until the exploration budget of $S$ steps is exhausted.

\subsection{Preliminary: 3D Gaussian Splatting}

We represent the scene as a set of anisotropic 3D Gaussian primitives
$\mathcal{G}=\{g_i\}_{i=1}^{N}$, where each Gaussian $g_i$ is parameterized by
its mean $\boldsymbol{\mu}_i \in \mathbb{R}^3$,
color $\mathbf{c}_i \in \mathbb{R}^3$,
opacity $\eta_i \in [0,1]$,
scale $\mathbf{S}_i \in \mathbb{R}^{3 \times 3}$,
and rotation $\mathbf{R}_i \in \mathrm{SO}(3)$.
Its covariance is given by
\begin{equation}
\boldsymbol{\Sigma}_i = \mathbf{R}_i \mathbf{S}_i \mathbf{S}_i^\top \mathbf{R}_i^\top.
\end{equation}
The rendered color is obtained via alpha compositing:
\begin{equation}
\hat{\mathcal{C}}(\mathbf{p}) =
\sum_{i=1}^{N} \alpha_i \mathbf{c}_i \prod_{j=1}^{i-1}(1-\alpha_j),
\end{equation}
where
\begin{equation}
\alpha_i =
\eta_i \exp\!\left(
-\tfrac{1}{2}
(\mathbf{p}-\boldsymbol{\mu}_i)^\top
\boldsymbol{\Sigma}_i^{-1}
(\mathbf{p}-\boldsymbol{\mu}_i)
\right).
\end{equation}
This formulation is fully differentiable and supports efficient online optimization.

\subsection{Dynamic-aware Gaussian Reconstruction}
\label{subsec:dyn_gaussian}

Dynamic environments violate the static-scene assumption: moving objects and transient occlusions introduce unreliable observations, which can corrupt Gaussian optimization and further mislead active planning. To address this issue, we develop an uncertainty-aware reconstruction strategy for dynamic active reconstruction. Specifically, we predict a per-pixel uncertainty map online, use it to suppress dynamically corrupted observations during Gaussian map optimization, and further lift it into map-level structural and motion uncertainty fields for downstream planning.

\subsubsection{Frame-level uncertainty prediction.}
For each input frame $I_t$, we extract dense visual features using a pre-trained DINOv3 feature extractor~\cite{oquab2023dinov2,simeoni2025dinov3}:
\begin{equation}
F_t = \mathcal{F}(I_t),
\end{equation}
where $\mathcal{F}$ denotes the DINOv3 encoder and $F_t$ is the resulting feature map.
These features are fed into a lightweight uncertainty prediction network $\mathcal{P}$ to estimate a per-pixel uncertainty map
\begin{equation}
\beta_t = \mathcal{P}(F_t),
\label{eq:beta_pred}
\end{equation}
where $\beta_t \in \mathbb{R}^{H\times W}$ is bilinearly upsampled to the input resolution.
Here, $\beta_t(\mathbf{u})$ measures the reliability of pixel $\mathbf{u}$, and pixels affected by dynamic motion or transient occlusion tend to have higher uncertainty values.
The uncertainty predictor is supervised by appearance and depth consistency; the full objective is provided in the supplementary material.

\subsubsection{Uncertainty-weighted Gaussian mapping.}
Let $\mathcal{G}_t$ denote the current Gaussian map.
Given the camera pose $\mathbf{T}_t$ and intrinsics $\mathbf{K}$, we render the predicted RGB image $\hat{I}_t$ and depth map $\hat{D}_t$ from $\mathcal{G}_t$.
The Gaussian map is optimized using an uncertainty-weighted rendering loss
\begin{equation}
\mathcal{L}_{\text{render}}
=
\frac{\lambda_1 \mathcal{L}_{\text{color}} + \lambda_2 \mathcal{L}_{\text{depth}}}{\beta_t^2}
+
\lambda_3 \mathcal{L}_{\text{iso}},
\label{eq:render_loss_dyn}
\end{equation}
where $\mathcal{L}_{\text{color}}$ combines pixel-wise $\ell_1$ and SSIM losses, and $\mathcal{L}_{\text{iso}}$ denotes isotropic Gaussian regularization.
Here, the uncertainty map $\beta_t$ acts as a pixel-wise weighting factor, so that unreliable observations contribute less to map optimization~\cite{ren2024nerf,zheng2025wildgs,Li2026DROIDW}.
This encourages the optimizer to focus on stable static structures while suppressing dynamic distractors during Gaussian updates.

\subsubsection{Map-level structural and motion uncertainty.}
To support dynamic-aware planning, we maintain two map-level uncertainty fields over a discrete 3D query set $\Omega$:
\begin{equation}
U_s:\Omega\rightarrow \mathbb{R}_{\ge 0}, \qquad
U_m:\Omega\rightarrow \mathbb{R}_{\ge 0},
\label{eq:uncertainty_fields}
\end{equation}
where $U_s(\mathbf{x})$ measures structural uncertainty and $U_m(\mathbf{x})$ measures motion-induced unreliability at location $\mathbf{x}$.

We estimate structural uncertainty from both insufficient observation and reconstruction mismatch. For each $\mathbf{x}\in\Omega$, let $n_t(\mathbf{x})$ denote the effective observation count and let $\bar e_t(\mathbf{x})$ denote an aggregated rendering residual. We define
\begin{equation}
U_s(\mathbf{x})
=
\lambda_c \frac{1}{\sqrt{n_t(\mathbf{x})+\epsilon}}
+
\lambda_r \,\mathrm{clip}\!\left(\bar e_t(\mathbf{x}),0,e_{\max}\right),
\label{eq:struct_uncertainty}
\end{equation}
where the first term highlights under-observed regions and the second term emphasizes locations that are poorly explained by the current Gaussian map.

Motion uncertainty is obtained by lifting the frame-level uncertainty map $\beta_t$ into 3D and accumulating it over time. For each query location $\mathbf{x}\in\Omega$, let $\mathbf{u}_t=\Pi(\mathbf{T}_t,\mathbf{x})$ denote its projection into frame $t$, and let $\mathbf{1}_t(\mathbf{x})$ indicate whether it is observable in that frame. We update motion uncertainty using
\begin{equation}
U_m(\mathbf{x}) \leftarrow (1-\alpha)\,U_m(\mathbf{x})
+
\alpha\,\mathbf{1}_t(\mathbf{x})\,\beta_t(\mathbf{u}_t),
\label{eq:motion_uncertainty}
\end{equation}
where $\alpha\in(0,1)$ controls the update rate.
As a result, $U_s$ identifies where more static evidence is needed, while $U_m$ identifies where future observations are likely to be unreliable due to dynamic disturbances.

\subsection{Dynamic-aware Viewpoint Selection}
\label{subsec:active_view}

Given the map-level structural uncertainty $U_s$ and motion uncertainty $U_m$, we select viewpoints that are both informative and dynamically reliable.
To improve long-horizon exploration efficiency in multi-room environments, we adopt a hierarchical planning strategy on a Voronoi graph~\cite{wu2024voronav}
$G=(\mathcal{V},\mathcal{E})$, where each node $n\in\mathcal{V}$ denotes a reachable region and each edge encodes traversal connectivity.

\subsubsection{Dynamic-aware subregion partition.}
Rather than planning over all nodes uniformly, we partition the graph into subregions using agglomerative clustering.
To account for dynamic disturbances, the pairwise distance between nodes $n_i$ and $n_j$ is defined as
\begin{equation}
\label{eq:dyn_cluster_dist}
d_{\mathrm{dyn}}(n_i,n_j)
=
\lambda_e d_E(n_i,n_j)
+
\lambda_p d_P(n_i,n_j)
+
\lambda_m d_M(n_i,n_j),
\end{equation}
where $d_E$ is the Euclidean distance, $d_P$ is the graph travel distance, and $d_M$ is the accumulated motion risk along the connecting path.
This produces a set of subregions $\{R_k\}$, including the current local subregion $R_l$ containing the agent.

\subsubsection{Local-global viewpoint selection.}
We prioritize local exploration before switching to global exploration.
For each candidate node $n\in R_l$, we define a local score
\begin{equation}
\label{eq:local_score}
S_{\mathrm{local}}(n)
=
\mathcal{I}_s(n;U_s)
-
\lambda_r \mathcal{R}_m(n;U_m)
-
\lambda_c \mathcal{C}(n),
\end{equation}
where $\mathcal{C}(n)$ is the travel cost from the current pose to node $n$.
The structural utility term is
\begin{equation}
\label{eq:node_gain}
\mathcal{I}_s(n;U_s)
=
\sum_{\mathbf{x}\in\Omega}
\mathrm{Vis}(n,\mathbf{x})\,U_s(\mathbf{x}),
\end{equation}
and the motion risk term is
\begin{equation}
\label{eq:node_risk}
\mathcal{R}_m(n;U_m)
=
\sum_{\mathbf{x}\in\Omega}
\mathrm{Vis}(n,\mathbf{x})\,U_m(\mathbf{x}),
\end{equation}
where $\mathrm{Vis}(n,\mathbf{x})\in[0,1]$ is a soft visibility score under the current Gaussian map $\mathcal{G}_t$.

The best local node is selected iteratively as long as its score remains above a threshold.
When the local region is either sufficiently explored or becomes unstable due to high motion uncertainty, the planner switches to global selection.
For each candidate subregion $R_k\neq R_l$, we define
\begin{equation}
\label{eq:global_region_score}
\begin{split}
S_{\mathrm{global}}(R_k)
&=
\sum_{n\in R_k}
\bigl(\mathcal{I}_s(n;U_s)-\lambda_r \mathcal{R}_m(n;U_m)\bigr) \\
&\quad
-\lambda_d\, d(R_l,R_k)
-\lambda_v\,P_{\mathrm{visit}}(R_k).
\end{split}
\end{equation}
where $d(R_l,R_k)$ is the inter-region travel cost and $P_{\mathrm{visit}}(R_k)$ penalizes repeatedly visited regions.
The region with the highest score is selected as the next exploration target, and a local candidate view action $a_t^\star=(\mathbf{p}_t^\star,\boldsymbol{\theta}_t^\star)$ is chosen around the selected node.

\subsection{Dynamic-constrained Path Planning}
\label{subsec:dynamic_plan}

After selecting the next target viewpoint, we compute a feasible path that jointly accounts for geometry and dynamic risk.
Unlike static planning, the objective here is not simply to minimize travel distance, but also to avoid routes likely to produce unstable or motion-corrupted observations.

For each graph edge $e\in\mathcal{E}$, we define a dynamic-constrained traversal cost
\begin{equation}
\label{eq:edge_cost_dyn}
c(e)=\ell(e)+\lambda_m \bar{U}_m(e),
\end{equation}
where $\ell(e)$ is the geometric edge length and $\bar{U}_m(e)$ is the average motion uncertainty along the edge.
The cost of a path $\pi$ is then
\begin{equation}
\label{eq:path_cost_dyn}
\mathcal{C}_{\mathrm{path}}(\pi)=\sum_{e\in\pi} c(e).
\end{equation}
This encourages the robot to favor dynamically stable routes instead of purely shortest paths.

To improve temporal robustness, motion uncertainty is accumulated over time using an exponential moving average:
\begin{equation}
\label{eq:motion_temporal}
U_m^{(t)}(\mathbf{x})
=
(1-\alpha)\,U_m^{(t-1)}(\mathbf{x})
+
\alpha\,\beta_t(\mathbf{x}),
\end{equation}
where $\beta_t$ is the frame-level uncertainty map predicted in Sec.~\ref{subsec:dyn_gaussian}.
As a result, transient disturbances do not dominate planning, while persistently dynamic regions are consistently down-weighted.

By combining dynamic-aware viewpoint selection with motion-constrained path planning, DynActiveGS forms a closed-loop active reconstruction system.
At each exploration step, the robot updates the Gaussian map and uncertainty fields from the latest RGB-D observation, selects the next informative target, and executes a dynamically stable path toward it.
The loop continues until the exploration step budget $S$ is exhausted, as summarized in the supplementary material Algorithm~\ref{alg:DynActiveGS}.

\section{Experiments}

\begin{figure*}[t]
  \centering
   \includegraphics[width=1.0\linewidth]{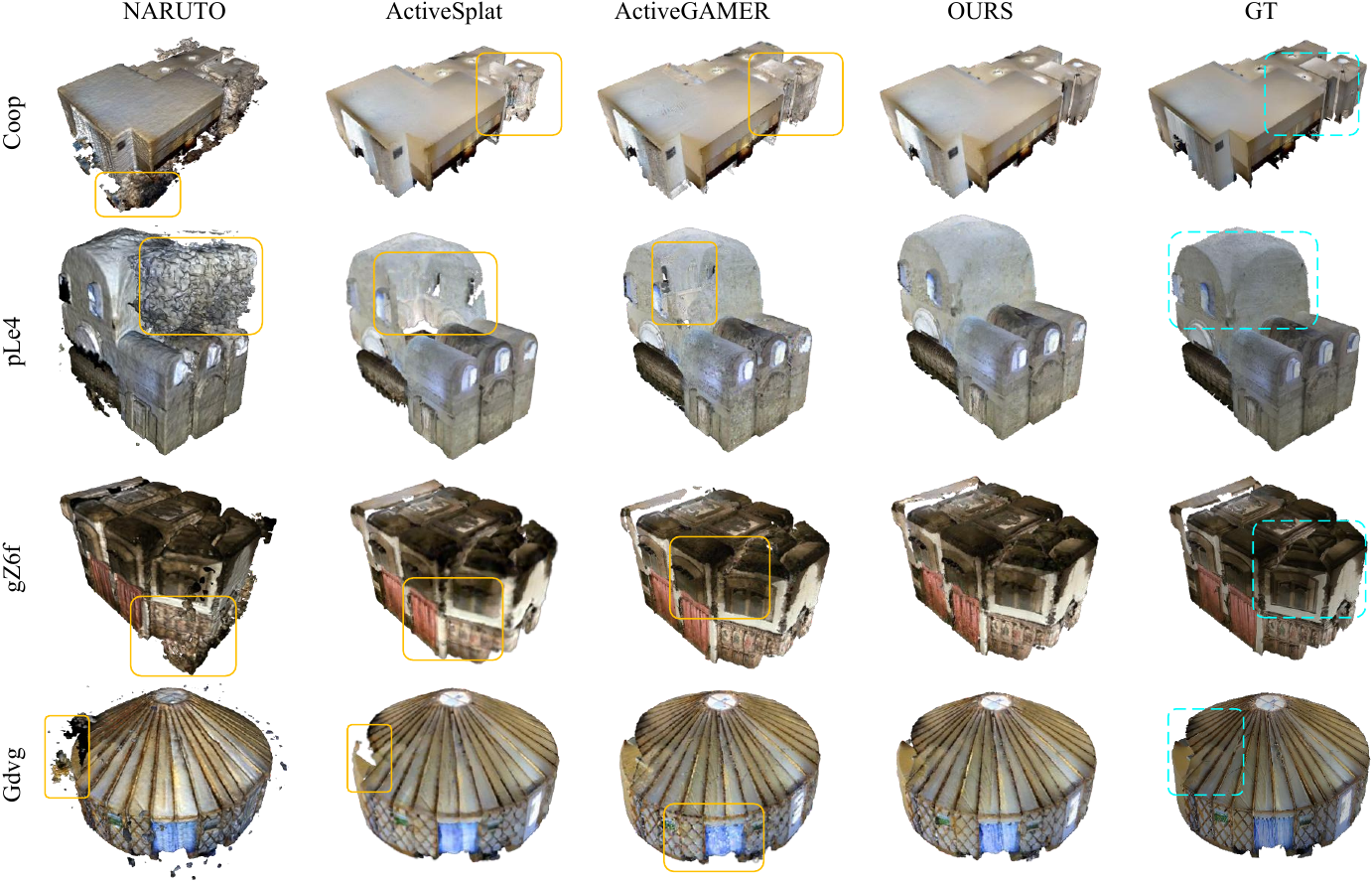}
\caption{Qualitative comparison of 3D reconstruction results on representative scenes from Social-MP3D and Dynamic Gibson. Blue bounding boxes indicate reference areas for easier comparison, while orange ones highlight low-quality reconstruction.}
   \label{fig:exp:qualitative_3d}
\end{figure*}

\subsection{Experimental Setup}

\begin{figure*}[t]
  \centering
   \includegraphics[width=1.0\linewidth]{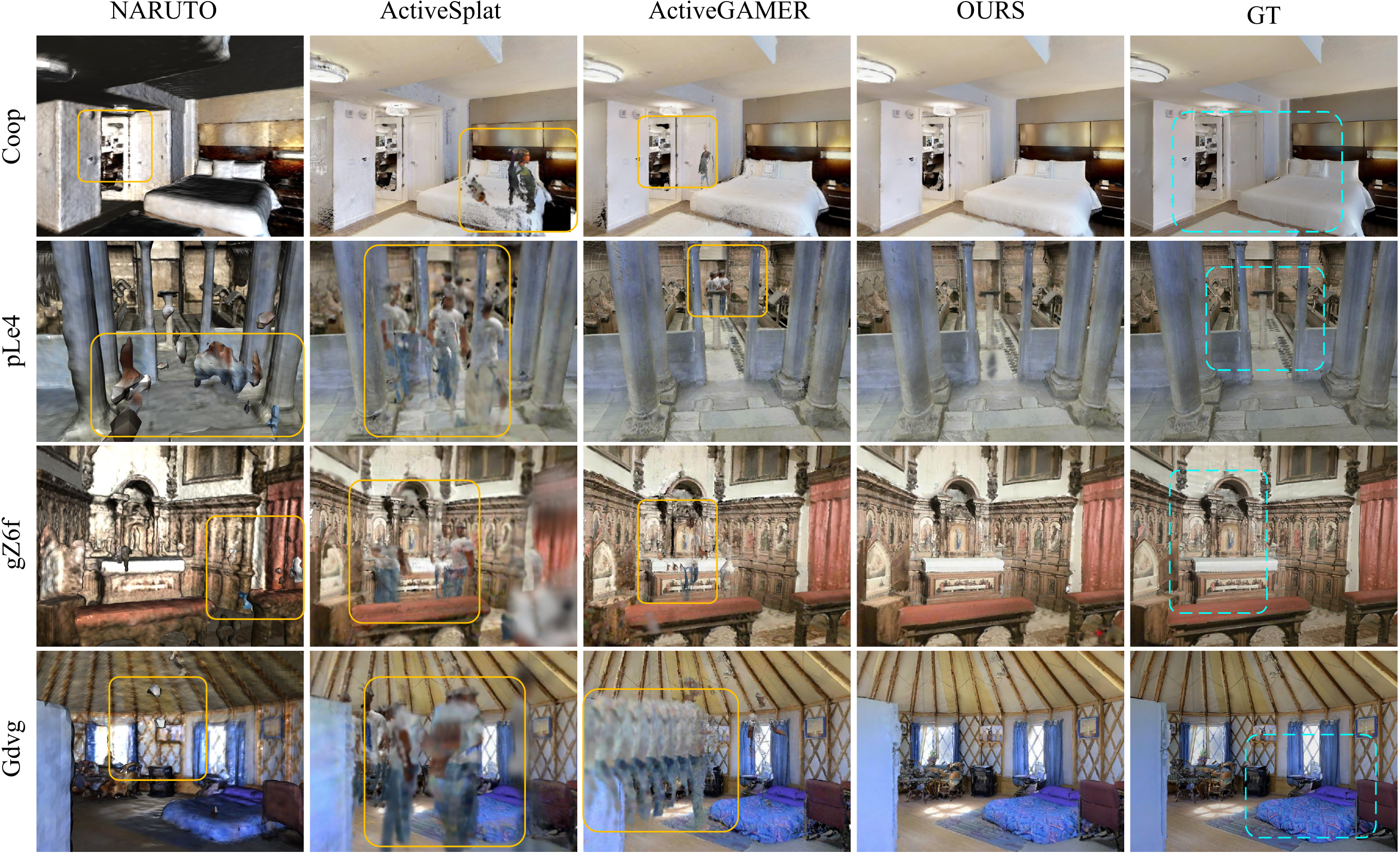}
\caption{\textbf{Novel view synthesis results on representative scenes from Social-MP3D and Dynamic Gibson.} The evaluated viewpoints are held-out views that do not appear in the training trajectories of any compared method. Blue bounding boxes indicate reference regions for easier comparison, while orange bounding boxes highlight low-quality renderings.}
   \label{fig:exp:qualitative_2d}
\end{figure*}

\textbf{Simulator and Dynamic Scenes.}
All experiments are conducted in the Habitat simulator~\cite{savva2019habitat}.
To evaluate dynamic active reconstruction, we use indoor scenes with runtime human dynamics in two settings.
For Matterport3D (MP3D)~\cite{chang2018matterport3d}, we adopt the public Social-MP3D benchmark~\cite{gong2025cognition}, which augments MP3D scenes with dynamic humans.
For Gibson~\cite{xia2018gibson}, since no public benchmark provides runtime humans for active reconstruction, we construct a reproducible \emph{Dynamic Gibson Protocol} in Habitat-Sim, following the human simulation paradigm of HabiCrowd~\cite{vuong2024habicrowd}.
In both settings, virtual pedestrians are instantiated at runtime with randomized start--goal waypoints and simulated using ORCA~\cite{van2011reciprocal} and UPL++~\cite{karamouzas2014universal,vuong2024habicrowd} crowd dynamics.
The dynamic humans are not part of the static scene geometry used for reconstruction evaluation.
We use fixed scene splits, shared human configurations, and standardized random seeds across all compared methods.

\noindent \textbf{Baselines.}
We compare DynActiveGS against state-of-the-art active reconstruction methods, including 3DGS-based approaches (ActiveGAMER~\cite{chen2025activegamer}, ActiveGS~\cite{jin2025activegs}, ActiveSplat~\cite{li_activesplat_2025}) and neural active mapping methods (NARUTO~\cite{feng2024naruto}).
Although some baselines employ free-flying 6DoF cameras, we evaluate them using their official implementations under identical observation budgets.
We also include dynamic passive mapping baselines without next-best-view planning.

\noindent \textbf{Geometric Metrics.}
We evaluate geometric reconstruction using Accuracy (Acc), Completion (Com), and Completion Ratio (C.R.) with a 5\,cm threshold.
Metrics are computed by uniformly sampling points from static ground-truth meshes and comparing them with point clouds extracted from reconstructed Gaussian maps.
Since dynamic humans are instantiated at runtime and are excluded from the ground-truth meshes, these metrics measure robustness of static scene reconstruction under dynamic disturbances.

\noindent \textbf{Rendering Metrics.}
For novel view rendering evaluation, we follow the protocol of the baseline ActiveGAMER~\cite{chen2025activegamer} and use same predefined novel-view trajectories for testing.
Specifically, we compare the rendered RGB images against static ground-truth renderings at identical poses, and report PSNR, SSIM, and LPIPS.
For depth rendering performance, we additionally use Depth L1 distance as the evaluation metric.

\noindent \textbf{Implementation Details.}
We consider a realistic ground mobile robot with a fixed camera height.
At each step, the agent executes one discrete navigation action and acquires one posed RGB-D observation along the planned viewpoint.
The camera field of view is set to $60^\circ$ vertically and $90^\circ$ horizontally, and the system processes image sequences online with on-policy planning and incremental reconstruction.
Unlike prior works that assume free-flying 6DoF cameras, our embodiment better reflects practical ground robot constraints.
All methods are evaluated under an equal observation budget: each method is allowed 1000 exploration steps, and each step produces exactly one RGB-D frame, ensuring identical input image counts across all approaches.
\label{sec:experiments}

\subsection{Comparison Results}

\textbf{Results on Social-MP3D.}
As shown in Tab.~\ref{tab:mp3d_dynamic}, the advantage of DynActiveGS becomes even more evident on the larger and more heavily occluded Social-MP3D scenes.
Under the same observation budget, static-scene baselines suffer substantial performance degradation in complex layouts with frequent human occlusions, whereas our method maintains stable reconstruction quality.

\begin{table}[t]
\caption{Quantitative comparison on the Social-MP3D dataset for 3D reconstruction and novel view synthesis.}
\centering
\scriptsize
\setlength{\tabcolsep}{2.5pt}
\resizebox{\columnwidth}{!}{
\begin{tabular}{lccccccc}
\toprule
\textbf{Method} & \textbf{Metric} & \textbf{Gdvg} & \textbf{gZ6f} & \textbf{HxpK} & \textbf{pLe4} & \textbf{YmJk} & \textbf{Avg.} \\

\midrule

\multirow{6}{*}{NARUTO~\cite{feng2024naruto}}
& Acc (cm)$\downarrow$  & 6.62 & 5.65 & 10.87 & 6.35 & 13.02 & 8.50 \\
& Com. (cm)$\downarrow$ & 7.91 & 4.26 & 9.41 & 5.98 & 9.73 & 7.46 \\
& C.R. (\%)$\uparrow$   & 82.46 & 85.11 & 83.27 & 79.42 & 76.76 & 81.40 \\
& PSNR$\uparrow$        & 18.63 & 18.51 & 15.46 & 20.24 & 16.45 & 17.86 \\
& SSIM$\uparrow$        & 0.624 & 0.548 & 0.432 & 0.697 & 0.386 & 0.537 \\
& LPIPS$\downarrow$     & 0.431 & 0.453 & 0.561 & 0.482 & 0.496 & 0.485 \\

\midrule

\multirow{6}{*}{ActiveSplat~\cite{li_activesplat_2025}}
& Acc (cm)$\downarrow$  & 4.64 & 3.05 & 5.98 & 6.74 & 9.27 & 5.94 \\
& Com. (cm)$\downarrow$ & 4.72 & 3.48 & 6.75 & 4.22 & 5.91 & 5.02 \\
& C.R. (\%)$\uparrow$   & 88.27 & 87.16 & 85.54 & 86.64 & 85.22 & 86.57 \\
& PSNR$\uparrow$        & 19.65 & 19.12 & 21.85 & 21.57 & 20.47 & 20.53 \\
& SSIM$\uparrow$        & 0.662 & 0.646 & 0.788 & 0.763 & 0.655 & 0.703 \\
& LPIPS$\downarrow$     & 0.516 & 0.552 & 0.387 & 0.423 & 0.458 & 0.467 \\

\midrule

\multirow{6}{*}{ActiveGAMER~\cite{chen2025activegamer}}
& Acc (cm)$\downarrow$  & 3.71 & 3.37 & 3.78 & 5.43 & 5.42 & 4.34 \\
& Com. (cm)$\downarrow$ & 5.76 & 3.91 & 4.11 & 6.08 & 6.64 & 5.30 \\
& C.R. (\%)$\uparrow$   & 92.18 & 92.74 & 91.46 & 88.12 & 86.35 & 90.17 \\
& PSNR$\uparrow$        & 19.76 & 20.23 & 22.83 & 23.07 & 21.52 & 21.48 \\
& SSIM$\uparrow$        & 0.670 & 0.693 & 0.816 & 0.826 & 0.721 & 0.745 \\
& LPIPS$\downarrow$     & 0.445 & 0.420 & 0.372 & 0.425 & 0.386 & 0.410 \\

\midrule

\multirow{6}{*}{\textbf{Ours}}
& Acc (cm)$\downarrow$  & \textbf{2.54} & \textbf{2.11} & \textbf{3.06} & \textbf{2.62} & \textbf{3.97} & \textbf{2.86} \\
& Com. (cm)$\downarrow$ & \textbf{2.87} & \textbf{2.46} & \textbf{3.82} & \textbf{3.02} & \textbf{4.28} & \textbf{3.29} \\
& C.R. (\%)$\uparrow$   & \textbf{94.41} & \textbf{93.02} & \textbf{95.58} & \textbf{94.77} & \textbf{92.63} & \textbf{94.08} \\
& PSNR$\uparrow$        & \textbf{21.36} & \textbf{20.74} & \textbf{26.23} & \textbf{24.98} & \textbf{21.87} & \textbf{23.04} \\
& SSIM$\uparrow$        & \textbf{0.722} & \textbf{0.703} & \textbf{0.894} & \textbf{0.862} & \textbf{0.755} & \textbf{0.787} \\
& LPIPS$\downarrow$     & \textbf{0.316} & \textbf{0.338} & \textbf{0.185} & \textbf{0.362} & \textbf{0.196} & \textbf{0.279} \\

\bottomrule
\end{tabular}}
\label{tab:mp3d_dynamic}
\end{table}

\begin{table}[t]
\caption{Quantitative comparison on the Dynamic Gibson for 3D reconstruction and novel view synthesis.}
\centering
\scriptsize
\setlength{\tabcolsep}{2.3pt}
\resizebox{\columnwidth}{!}{
\begin{tabular}{lcccccccccc}
\toprule
\textbf{Method} & \textbf{Metric} & \textbf{Beac} & \textbf{Coop} & \textbf{Denm} & \textbf{Elmi} & \textbf{Eudo} & \textbf{Grei} & \textbf{Home} & \textbf{Ribe} & \textbf{Avg.} \\
\midrule

\multirow{6}{*}{NARUTO~\cite{feng2024naruto}}
& Acc (cm)$\downarrow$  & 5.42 & 4.75 & 4.62 & 6.31 & 5.14 & 5.26 & 4.58 & 4.38 & 5.06 \\
& Com. (cm)$\downarrow$ & 5.72 & 5.36 & 5.18 & 7.02 & 5.72 & 5.88 & 5.12 & 4.95 & 5.66 \\
& C.R. (\%)$\uparrow$   & 84.38 & 86.95 & 87.42 & 81.64 & 85.91 & 84.72 & 88.62 & 88.76 & 86.05 \\
& PSNR$\uparrow$        & 18.94 & 18.68 & 17.91 & 16.85 & 18.36 & 16.97 & 19.31 & 20.82 & 18.48 \\
& SSIM$\uparrow$        & 0.691 & 0.737 & 0.682 & 0.659 & 0.676 & 0.586 & 0.687 & 0.763 & 0.685 \\
& LPIPS$\downarrow$     & 0.472 & 0.449 & 0.468 & 0.496 & 0.531 & 0.488 & 0.469 & 0.456 & 0.479 \\

\midrule

\multirow{6}{*}{ActiveSplat~\cite{li_activesplat_2025}}
& Acc (cm)$\downarrow$  & 5.61 & 4.52 & 4.05 & 4.78 & 4.18 & 4.65 & 3.96 & 3.87 & 4.45 \\
& Com. (cm)$\downarrow$ & 6.25 & 5.03 & 4.58 & 5.41 & 4.72 & 5.16 & 4.34 & 4.26 & 4.97 \\
& C.R. (\%)$\uparrow$   & 84.26 & 88.37 & 89.94 & 86.91 & 89.18 & 87.05 & 90.36 & 90.82 & 88.36 \\
& PSNR$\uparrow$        & 22.22 & 21.05 & 19.34 & 20.17 & 20.88 & 22.43 & 21.81 & 20.72 & 21.08 \\
& SSIM$\uparrow$        & 0.797 & 0.833 & 0.789 & 0.714 & 0.761 & 0.736 & 0.731 & 0.782 & 0.768 \\
& LPIPS$\downarrow$     & 0.323 & 0.369 & 0.519 & 0.445 & 0.531 & 0.339 & 0.456 & 0.451 & 0.429 \\

\midrule

\multirow{6}{*}{ActiveGAMER~\cite{chen2025activegamer}}
& Acc (cm)$\downarrow$  & 5.08 & 3.72 & 4.14 & 4.31 & 4.08 & 3.61 & 3.51 & 3.42 & 3.98 \\
& Com. (cm)$\downarrow$ & 5.74 & 4.28 & 4.71 & 4.92 & 4.56 & 4.15 & 3.96 & 3.88 & 4.53 \\
& C.R. (\%)$\uparrow$   & 87.84 & 92.03 & 90.47 & 90.12 & 91.35 & 92.76 & 93.11 & 93.48 & 91.40 \\
& PSNR$\uparrow$        & 23.47 & 22.11 & 20.42 & 21.34 & 21.96 & 22.58 & 23.95 & 22.08 & 22.24 \\
& SSIM$\uparrow$        & 0.794 & 0.832 & 0.747 & 0.723 & 0.756 & 0.729 & 0.748 & 0.785 & 0.764 \\
& LPIPS$\downarrow$     & 0.338 & 0.391 & 0.486 & 0.414 & 0.398 & 0.366 & 0.409 & 0.372 & 0.397 \\

\midrule

\multirow{6}{*}{\textbf{Ours}}
& Acc (cm)$\downarrow$  & \textbf{2.14} & \textbf{2.31} & \textbf{2.24} & \textbf{2.68} & \textbf{3.52} & \textbf{2.76} & \textbf{3.45} & \textbf{2.03} & \textbf{2.52} \\
& Com. (cm)$\downarrow$ & \textbf{2.56} & \textbf{2.78} & \textbf{2.71} & \textbf{3.26} & \textbf{3.04} & \textbf{3.35} & \textbf{3.98} & \textbf{2.48} & \textbf{3.02} \\
& C.R. (\%)$\uparrow$   & \textbf{96.18} & \textbf{95.37} & \textbf{95.84} & \textbf{93.86} & \textbf{94.92} & \textbf{93.94} & \textbf{91.75} & \textbf{96.73} & \textbf{94.82} \\
& PSNR$\uparrow$        & \textbf{25.28} & \textbf{24.34} & \textbf{21.86} & \textbf{23.91} & \textbf{24.18} & \textbf{24.86} & \textbf{25.65} & \textbf{24.62} & \textbf{24.34} \\
& SSIM$\uparrow$        & \textbf{0.866} & \textbf{0.858} & \textbf{0.811} & \textbf{0.771} & \textbf{0.842} & \textbf{0.835} & \textbf{0.841} & \textbf{0.852} & \textbf{0.835} \\
& LPIPS$\downarrow$     & \textbf{0.254} & \textbf{0.243} & \textbf{0.339} & \textbf{0.262} & \textbf{0.298} & \textbf{0.286} & \textbf{0.311} & \textbf{0.307} & \textbf{0.288} \\

\bottomrule
\end{tabular}}
\label{tab:gibson_dynamic}
\end{table}

\noindent \textbf{Results on Dynamic Gibson.}
Tab.~\ref{tab:gibson_dynamic} reports the quantitative results on Gibson scenes augmented with our pedestrian injection protocol.
Although Gibson has traditionally been used as a static active reconstruction benchmark, introducing dynamic humans leads to frequent occlusions and motion-corrupted observations, which pose significant challenges to next-best-view planning.

DynActiveGS consistently outperforms all baselines in both geometric and photometric metrics.
Compared with the strongest static-scene active reconstruction baseline, our method reduces reconstruction error and improves completion ratio.
More importantly, the rendering quality is substantially improved, as reflected by higher PSNR and significantly lower LPIPS.

These results demonstrate that the proposed motion-aware uncertainty modeling generalizes effectively to active reconstruction in dynamic scenes.
While static-scene baselines continue to expand coverage, they cannot filter motion-contaminated observations, often resulting in duplicated structures and texture inconsistencies.
By contrast, DynActiveGS explicitly disentangles structural uncertainty from motion-induced uncertainty, enabling the agent to avoid high-risk viewpoints and maintain stable map updates.

More importantly, this experiment shows that the gains of our method are not tied to a specific dataset.
Even when dynamic agents are introduced through an external augmentation protocol rather than being natively embedded in the dataset, our framework still exhibits strong robustness and consistent reconstruction quality.

\begin{figure}[t]
    \centering
    \includegraphics[width=\columnwidth]{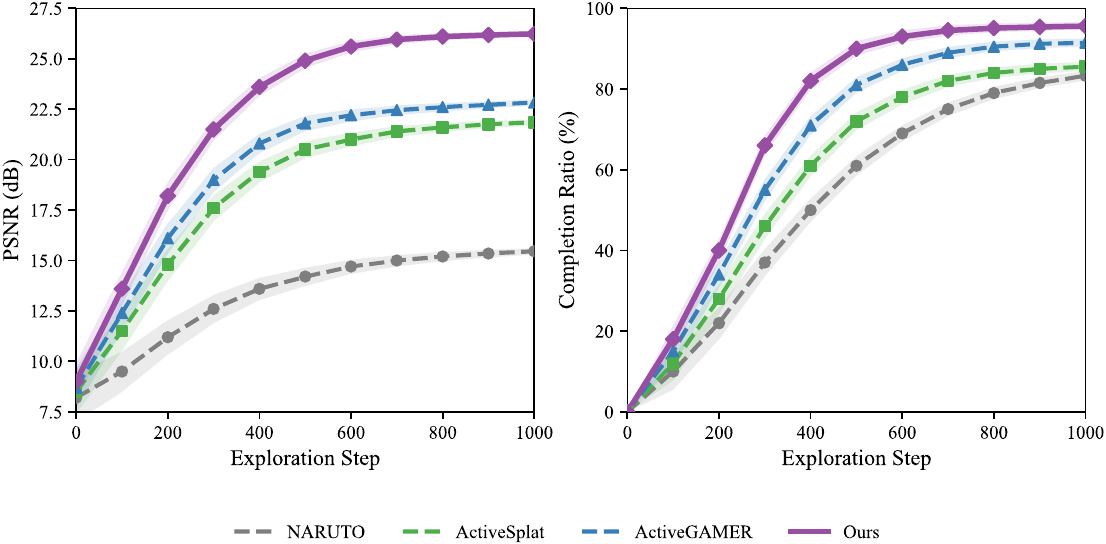}
\caption{Reconstruction progress on Social-MP3D \textit{HxpK}. DynActiveGS achieves the fastest improvement in both PSNR and completion ratio throughout the exploration process.}
    \label{fig:progress_hxpk}
\end{figure}

\begin{figure}[t]
    \centering
    \includegraphics[width=\columnwidth]{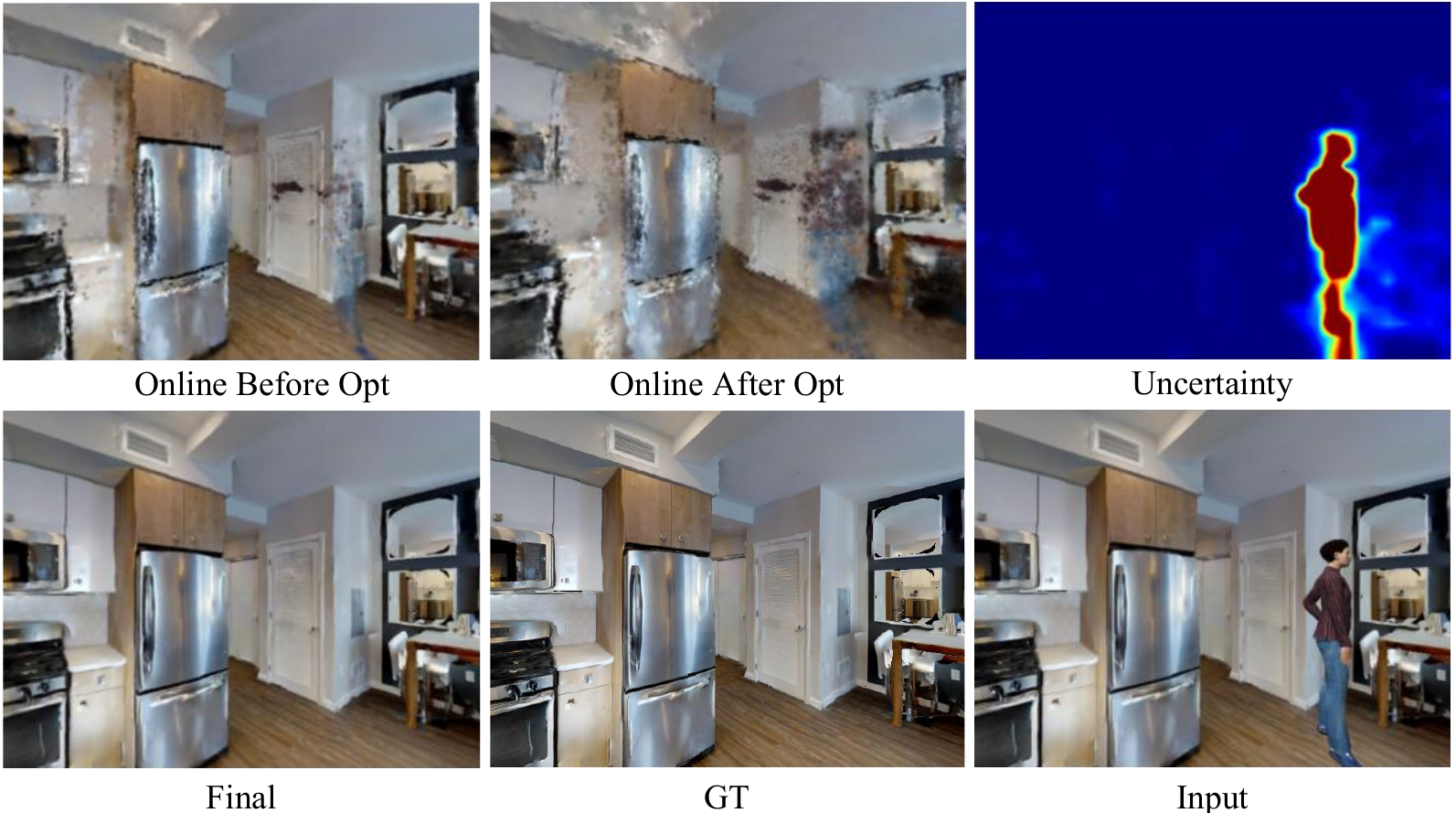}
\caption{Rendering quality across reconstruction stages on Dynamic Gibson \textit{Denm} scene. The proposed uncertainty-aware optimization progressively suppresses dynamic artifacts and improves reconstruction quality from the initial online result to the final refined result.}
    \label{fig:stagewise_render}
\end{figure}

\begin{table*}[t]
  \vspace{-6pt}
  \centering
  \footnotesize
  \renewcommand{\arraystretch}{0.9}
  \caption{
    \textbf{Ablation of core components.}
    We progressively enable uncertainty-weighted Gaussian mapping, motion-aware viewpoint selection, and dynamic-constrained planning.
    Each component brings consistent gains, and the full model achieves the best balance between geometric reconstruction and rendering quality.
  }
  \begin{tabular*}{\textwidth}{@{\extracolsep{\fill}} lccc|ccccccc @{}}
    \toprule
    Variant & Weighted Map & Motion-aware View & Dyn. Planner &
    Acc$\downarrow$ & Com$\downarrow$ & C.R.$\uparrow$ &
    PSNR$\uparrow$ & SSIM$\uparrow$ & LPIPS$\downarrow$ & L1-D$\downarrow$ \\
    \midrule
    Baseline & \ding{55} & \ding{55} & \ding{55}
    & 3.21 & 3.84 & 90.62 & 18.71 & 0.503 & 0.505 & 6.94 \\

    + Weighted Mapping & \ding{51} & \ding{55} & \ding{55}
    & 2.98 & 3.56 & 91.85 & 19.83 & 0.621 & 0.471 & 5.12 \\

    + Motion-aware View & \ding{51} & \ding{51} & \ding{55}
    & 2.76 & 3.28 & 92.93 & 20.66 & 0.733 & 0.409 & 4.08 \\

    \midrule
    \textbf{Full (Ours)} & \ding{51} & \ding{51} & \ding{51}
    & \textbf{2.24} & \textbf{2.71} & \textbf{95.84}
    & \textbf{21.86} & \textbf{0.811} & \textbf{0.339} & \textbf{2.92} \\
    \bottomrule
  \end{tabular*}
  \label{tab:ablation_core}
  \vspace{-6pt}
\end{table*}

\begin{table}[t]
  \vspace{-4pt}
  \centering
  \footnotesize
  \renewcommand{\arraystretch}{0.9}
  \caption{
    \textbf{Ablation of motion-aware viewpoint selection.}
  }
  \begin{tabular*}{\columnwidth}{@{\extracolsep{\fill}} lccc|ccc @{}}
    \toprule
    Variant & $U_s$ & $U_m$ & Cost &
    Acc$\downarrow$ & C.R.$\uparrow$ & PSNR$\uparrow$ \\
    \midrule
    Random & \ding{55} & \ding{55} & \ding{55}
    & 3.08 & 91.24 & 19.37 \\

    Static-Gain & \ding{51} & \ding{55} & \ding{55}
    & 2.81 & 92.15 & 20.14 \\

    Gain+Cost & \ding{51} & \ding{55} & \ding{51}
    & 2.63 & 93.08 & 20.73 \\

    \textbf{Ours} & \ding{51} & \ding{51} & \ding{51}
    & \textbf{2.03} & \textbf{96.73} & \textbf{24.62} \\
    \bottomrule
  \end{tabular*}
  \label{tab:ablation_view}
  \vspace{-4pt}
\end{table}

\begin{table}[t]
\caption{
\textbf{Ablation of dynamic-constrained planning.}
}
\centering
\scriptsize
\setlength{\tabcolsep}{3.2pt}
\resizebox{\columnwidth}{!}{
\begin{tabular}{l|ccc|ccc|ccc}
\toprule
\multirow{2}{*}{Path Ratio} &
\multicolumn{3}{c|}{Static Hier.} &
\multicolumn{3}{c|}{+ Dyn. Edge Cost} &
\multicolumn{3}{c}{\textbf{Ours}} \\
& Com.$\downarrow$ & C.R.$\uparrow$ & PSNR$\uparrow$
& Com.$\downarrow$ & C.R.$\uparrow$ & PSNR$\uparrow$
& Com.$\downarrow$ & C.R.$\uparrow$ & PSNR$\uparrow$ \\
\midrule
25\%  & 6.12 & 88.47 & 18.95 & 5.86 & 89.73 & 19.42 & \textbf{5.54} & \textbf{90.68} & \textbf{20.91} \\
50\%  & 5.36 & 90.22 & 19.71 & 5.08 & 91.84 & 20.74 & \textbf{4.28} & \textbf{92.63} & \textbf{21.87} \\
75\%  & 4.25 & 91.67 & 20.82 & 3.96 & 92.31 & 21.26 & \textbf{3.61} & \textbf{93.46} & \textbf{22.72} \\
100\% & 3.89 & 92.28 & 21.97 & 3.66 & 93.02 & 22.63 & \textbf{2.96} & \textbf{94.41} & \textbf{24.17} \\
\bottomrule
\end{tabular}
}
\label{tab:ablation_plan}
\end{table}

\noindent \textbf{Qualitative Comparison.}
Fig.~\ref{fig:exp:qualitative_3d} and Fig.~\ref{fig:exp:qualitative_2d} present qualitative comparisons.
When humans move across the field of view, static baselines tend to produce duplicated structures and motion streaks.
DynActiveGS effectively suppresses these artifacts while preserving consistent geometry and texture fidelity.
These observations further validate the benefit of risk-aware planning under dynamic disturbances.
These results suggest that the primary limitation of conventional active reconstruction in dynamic environments is not insufficient view sampling capacity, but the lack of motion-aware decision-making.

\noindent \textbf{Reconstruction Progress over Exploration.}
As shown in Fig.~\ref{fig:progress_hxpk}, DynActiveGS exhibits the fastest improvement in both rendering quality and reconstruction completeness on the Social-MP3D \textit{HxpK} scene.
Compared with NARUTO~\cite{feng2024naruto}, ActiveSplat~\cite{li_activesplat_2025}, and ActiveGAMER~\cite{chen2025activegamer}, our method achieves higher PSNR and completion ratio across nearly all exploration stages, further validating the effectiveness of the proposed dynamic-aware active reconstruction pipeline.

\subsection{Ablation Study}
\label{subsec:ablation}

We conduct ablation studies on representative dynamic scenes from Social-MP3D and Dynamic Gibson to validate the key components of DynActiveGS.
Our analysis covers three aspects: (1) core module ablation, (2) motion-aware viewpoint selection, and (3) dynamic-constrained planning.

\noindent \textbf{Core module ablation.}
We first evaluate the contribution of the main modules, including uncertainty-weighted Gaussian mapping, motion-aware viewpoint selection, and dynamic-constrained planning, using the Dynamic Gibson \textit{Denm} scene.
As shown in Tab.~\ref{tab:ablation_core}, each component consistently improves both reconstruction and rendering quality.
In particular, uncertainty-weighted mapping suppresses motion-corrupted observations, motion-aware viewpoint selection improves observation quality, and dynamic-constrained planning further enhances exploration efficiency.
The full model achieves the best overall performance across all metrics.

\noindent \textbf{Ablation of motion-aware viewpoint selection.}
We compare random viewpoint sampling (\textit{Random}), structural-gain-only selection (\textit{Static-Gain}), structural gain with travel cost (\textit{Gain+Cost}), and our full motion-aware formulation (\textit{Ours}) on the Dynamic Gibson \textit{Ribe} scene.
Tab.~\ref{tab:ablation_view} shows that using structural uncertainty alone is insufficient in dynamic scenes, since the agent is often attracted to regions with high gain but low observation reliability.
Introducing travel cost improves efficiency, while explicitly modeling motion uncertainty further improves both reconstruction accuracy and rendering fidelity.
These results confirm the importance of motion-aware viewpoint utility under human disturbances.

\noindent \textbf{Ablation of dynamic-constrained planning.}
We further analyze the planner by comparing a static hierarchical planner (\textit{Static Hier.}), a planner with dynamic edge cost but without temporal accumulation (\textit{+ Dyn. Edge Cost}), and our full dynamic-constrained planner (\textit{Ours}).
To evaluate long-horizon efficiency, we extend the exploration horizon to 2000 steps on the Social-MP3D \textit{YmJk} scene and report results at different path ratios in Tab.~\ref{tab:ablation_plan}.
Hierarchical planning already improves long-horizon exploration over naive shortest-path execution, while dynamic edge cost reduces unstable routes through highly dynamic regions.
Temporal accumulation yields the best final performance, suggesting that persistent motion patterns are more informative for planning than instantaneous disturbances.

\noindent \textbf{Stage-wise rendering analysis.}
Fig.~\ref{fig:stagewise_render} visualizes reconstruction quality at different stages of the pipeline.
The initial online result (\textit{Online Before Opt}) still contains strong dynamic artifacts.
After uncertainty-guided online optimization (\textit{Online After Opt}), these artifacts are substantially reduced.
The final offline refinement further improves consistency and produces the best visual quality.
Together with the uncertainty visualization, this result shows that the learned uncertainty map effectively highlights motion-corrupted regions and supports robust Gaussian optimization.

Overall, these ablations consistently validate the design of DynActiveGS.
Uncertainty-weighted mapping improves reconstruction robustness, motion-aware viewpoint selection improves observation quality, and dynamic-constrained planning enhances long-horizon exploration efficiency.
Together, these components enable robust active reconstruction in dynamic environments.





\section{Conclusion and Future Work}
\label{sec:conclusion}

We introduced \textbf{DynActiveGS}, a dynamic-aware active reconstruction framework based on 3D Gaussian Splatting.
By disentangling structural and motion-induced uncertainty, and integrating dynamic-aware viewpoint selection with constrained path planning, our method enables robust closed-loop exploration and high-fidelity reconstruction in dynamic environments.
Experiments on Social-MP3D and dynamically augmented Gibson scenes demonstrate consistent improvements over existing baselines in reconstruction accuracy, completeness, and rendering quality.

\noindent \textbf{Limitations.}
Our evaluation is currently conducted in simulation with controllable human dynamics, while real-world environments may contain more diverse and unpredictable motions.
Moreover, the framework is not yet optimized for large-scale multi-floor environments with complex spatial structures.

\noindent \textbf{Future Work.}
Future directions include deployment on real robotic platforms, online motion prediction, and extending dynamic-aware active reconstruction toward 4D scene representations~\cite{lin2026movies} for modeling persistent scene dynamics.

\clearpage
\section{Acknowledgments}
This work was supported by the National Natural Science Foundation of China under Grant Nos. 62293545 and U21B6002, in part by the Major Key Project of PCL under Grant Nos. PCL2024A06 and PCL2025A10, and in part by the Shenzhen Science and Technology Program under Grant Nos. RCJC20210706091946001, ZDCY2025090-1104207008, and RCJC20231211085918010.
\balance   
\bibliographystyle{ACM-Reference-Format}
\bibliography{sigconf}

\clearpage

\end{document}